%% file: IEEE_paper_main.tex
\documentclass[letterpaper, conference, final]{IEEEtran}
\usepackage{bm}
\usepackage{diagbox} 
\usepackage{makecell}
\usepackage{array} 
\usepackage{multirow}
\usepackage{cite}
\usepackage{amsmath,amssymb,amsfonts}
\usepackage{algorithmic}
\usepackage{booktabs}
\usepackage{graphicx}
\usepackage{textcomp}
\usepackage{makecell}
\usepackage{xcolor}
\usepackage{multirow}
\usepackage{import}
\usepackage[nolist]{acronym}
\usepackage{siunitx}
\usepackage[utf8]{inputenc}
\usepackage[hidelinks]{hyperref}

\def\BibTeX{{\rm B\kern-.05em{\sc i\kern-.025em b}\kern-.08em
    T\kern-.1667em\lower.7ex\hbox{E}\kern-.125emX}}

\usepackage{xspace}

\makeatletter
\DeclareRobustCommand\onedot{\futurelet\@let@token\@onedot}
\def\@onedot{\ifx\@let@token.\else.\null\fi\xspace}

\def\etc{\emph{etc}\onedot}

\makeatother

\title{\LARGE \bf Hybrid Machine Learning for Articulation Angle Estimation of Truck-Semitrailer Combinations}
\begin{acronym}
\acro{adas}[ADAS]{advanced driver assistance system}
\acroplural{adas}[ADAS]{advanced driver assistance systems}
\acro{ood}[OOD]{out-of-distribution}
\acro{cms}[CMS]{camera-mirror-system}
\acro{rmse}[RMSE]{root mean squared error}
\acro{mae}[MAE]{mean absolute error}
\acro{cnn}[CNN]{convolutional neural networks}
\acro{ekf}[EKF]{extended Kalman filter}
\acro{mlp}[MLP]{multilayer perceptron}
\acro{vslam}[VSLAM]{visual simultaneous localization and mapping}
\acro{vkit}[VKiT]{Visual Kinematic Transformer}
\acro{svkit}[SVKiT]{Sequential Visual Kinematic Transformer}
\acro{pca}[PCA]{Principal Component Analysis}
\end{acronym}

\begin{document}
\author{Qixuan Zhang$^{1,2}$, Jonas Boettcher$^{1}$, Simon F. G. Ehlers$^{2}$ and Marvin Stuede$^{1}$}

\begin{titlepage}
\thispagestyle{empty}

\vspace*{3cm}

\begin{center}

{\Large\textbf{Copyright Notice}}

\vspace{0.5cm}

\small
© 2026 IEEE.  Personal use of this material is permitted.  Permission from IEEE must be obtained for all other uses, in any current or future media, including reprinting/republishing this material for advertising or promotional purposes, creating new collective works, for resale or redistribution to servers or lists, or reuse of any copyrighted component of this work in other works.

\vspace{1.5cm}
Accepted to be published in: Proceedings of the 2026 International Conference on Intelligent Transportation Systems (ITSC) September 15 – 18, Naples, Italy

\vfill

\end{center}

\end{titlepage}

\maketitle
\renewcommand\thefootnote{} 
\footnotetext{\footnotesize
   $^{1}$Authors are with ZF CV Systems Hannover GmbH, D-30453 Hannover, Germany,
	       {\tt\small  \href{mailto:marvin.stuede@zf.com}{marvin.stuede@zf.com}}, {\tt\small \href{mailto:jonas.boettcher2@zf.com}{jonas.boettcher2@zf.com}}
    $^{2}$Authors are with the Leibniz University Hannover, Institute of Mechatronic Systems, D-30823 Garbsen, Germany,
	       {\tt\small \href{mailto:simon.ehlers@imes.uni-hannover.de}{simon.ehlers@imes.uni-hannover.de}}, {\tt\small \href{mailto:qixuan.zhang@stud.uni-hannover.de}{qixuan.zhang@stud.uni-hannover.de}}}
\begin{abstract}
Accurate articulation angle estimation of trucks with trailers is critical for autonomous driving and \acp{adas}. Existing methods either require manual initialization, additional sensors, or prior knowledge and signals from trailers, or they lack real-world validation, limiting practical deployment.
This paper presents multiple learning-based models to directly estimate articulation angles from visual and kinematic inputs, eliminating the need for dedicated driving maneuvers for initialization, bounding box annotations, trailer-mounted sensor signals, or prior knowledge of trailer parameters.
Two learning-based models are integrated with a kinematic model within an \ac{ekf} framework, and an adaptive weighting scheme based on uncertainty quantification is applied for measurements involving visual input. Extensive real-world experiments with different trailer types demonstrate the approaches' robustness and generalization under out-of-domain conditions, including new trailers, varying colors, and lighting conditions. Results show that the hybrid method achieves accurate and reliable articulation angle estimation while maintaining reduced implementation requirements and practical deployment advantages.
\end{abstract}


\section{Introduction}



Among the key state variables governing the dynamics of articulated vehicles, the articulation angle between a truck and its semi-trailer is of particular importance, as it directly affects vehicle maneuverability, lateral stability, and collision risk.
Therefore, accurate estimation of this angle is critical for numerous applications, including autonomous driving, trailer tracking, motion prediction, and active safety systems.
In \acp{adas} implementations, articulation angle estimation can be used to adapt the field-of-view of a \ac{cms} or to display the current configuration of the whole vehicle to the driver.
Such functionality enables early hazard warnings and enhances overall operational safety.


However, in most commercial truck platforms, direct measurement of the articulation angle is rarely available, as it requires dedicated hardware typically installed at the coupling point between the truck and the trailer.
Alternatively,  previous studies have estimated the articulation angle using additional sensors such as LIDAR or ultrasonic sensors \cite{korayem2021hitch,josef2016trailer,olutomilayo2019estimation,ljungqvist2019path,leng2016curvature}.
Such solutions increase system cost, installation complexity, and maintenance effort, thereby limiting their applicability in series-production vehicles.
Consequently, estimating the articulation angle using only sensors and signals already available on the truck, such as kinematic sensors and rear-facing mirror replacement cameras, presents a practical and scalable alternative that can be seamlessly integrated into existing vehicle architectures.
Moreover, performing the estimation without relying on additional trailer-mounted sensors or prior knowledge of the trailer configuration is essential for enabling rapid adaptation to previously unseen trailers.

Existing research on articulation angle estimation based on \emph{kinematic} inputs can roughly be divided into model-based, learning-based, and hybrid approaches.
Model-based methods formulate vehicle dynamics through linear or nonlinear tire models to estimate system states \cite{ziaukas2019simultaneous}.
While physically interpretable, this requires extensive parameter identification, which limits scalability to diverse truck–trailer configurations.
In contrast, learning-based methods map kinematic inputs directly to articulation angle estimates, often employing recurrent neural networks to exploit temporal dependencies \cite{han2022hybrid,jahn2020neural}.
However, their performance may degrade in \ac{ood} conditions \cite{jahn2020neural,ewering2024reliable}. 
To overcome these limitations, hybrid approaches integrate physical models with data-driven components.
Examples include systems combining lateral dynamic vehicle models with neural observers in \ac{ekf} frameworks \cite{ziaukas2019simultaneous,jahn2020neural}, or architectures coupling Kalman filters with gated recurrent unit networks for adaptive measurement fusion \cite{han2022hybrid}.
Such combinations further enable adaptive weighting of inputs based on the similarity between training and inference conditions \cite{ewering2024reliable}.
Physics-informed neural networks have also been applied to predict dynamic states continuously over time \cite{zeipel2024fast}, offering a balance between model fidelity and data adaptability.
However, purely kinematic-based methods often require an initialization procedure for the articulation angle, such as zero-initialization during a straight-driving phase, which limits their applicability when the vehicle starts with a large initial articulation angle.
These challenges motivate the development of more robust estimation methods capable of improving accuracy and generalization across operating conditions.

With the increasing adoption of \ac{cms} in modern trucks, rear-facing \emph{cameras} offer an appealing alternative by providing \emph{absolute} articulation angle measurements derived directly from visual information. Unlike kinematic models, camera-based approaches do not rely on initialization and can thus deliver immediate state estimates, improving robustness during start-up and under diverse operating conditions.
Traditional methods rely on geometric markers mounted on the trailer to infer the angle from camera pose \cite{liu2025vision,fuchs2017model,de2019measurement} or employ \ac{vslam} and template matching techniques \cite{de2021camera,de2015visual,de2019measurement}.
Deep learning-based vision methods have demonstrated improved robustness and generalization.
Convolutional neural network \acused{cnn} (\ac{cnn}) architectures have been applied to process mirror-mounted camera images from simulated and real datasets for direct articulation angle estimation \cite{thawainin2025surveying,dahal2019deeptrailerassist,bahramgiri2022hitch}, while models based on CNN and LSTM architectures are also used in related domains such as train–car coupling \cite{liu2024supervised}. 
Furthermore, several studies integrate image-based approaches with kinematic models within a Kalman filtering framework in order to enhance prediction accuracy \cite{de2015visual,de2019measurement}.

In summary, existing methods remain constrained by their reliance on additional sensors or external geometric markers mounted on the trailer \cite{korayem2021hitch,josef2016trailer,olutomilayo2019estimation,ljungqvist2019path,leng2016curvature,liu2024supervised,liu2025vision,fuchs2017model,de2019measurement}, the absence of initial articulation angle measurements \cite{ziaukas2019simultaneous,han2022hybrid,jahn2020neural,de2021camera,de2015visual,de2019measurement}, assumptions of planar trailer surfaces or the need for bounding box annotations \cite{dahal2019deeptrailerassist,bahramgiri2022hitch}, and limited real-world validation \cite{thawainin2025surveying}.
Motivated by these limitations, this paper addresses the stable and accurate online estimation of the articulation angle for truck–trailer combinations.
The proposed framework fuses camera data, captured within a typical \ac{cms} field of view, with kinematic measurements solely from onboard sensors of truck in a hybrid manner, enabling the usage with different semi-trailers.
Multiple learning-based and hybrid approaches are implemented and compared. The contributions of this paper are the following:
\begin{enumerate}
    \item Multiple \ac{cnn}-based and Transformer-based models are proposed to directly estimate the articulation angle from either pure visual input or fused visual and kinematic signals, eliminating the need for dedicated driving maneuvers for initialization, bounding box annotations, trailer-mounted sensor signals, or prior knowledge of trailer parameters.
    \item Applying for the first time an adaptive weighting scheme based on \cite{ewering2024reliable} to dynamically integrate visual estimation with a kinematic model within an \ac{ekf} framework for articulation angle estimation.
    \item Extensive real-world experiments are conducted with four different trailers, demonstrating the method’s generalization capabilities under out-of-domain conditions, including new trailer types, exterior colors, and lighting variations.
\end{enumerate}
\IEEEsettopmargin{t}{54pt}
Results demonstrate that the hybrid approaches achieve accurate articulation angle estimation while significantly enhancing robustness and generalization under out-of-domain conditions compared with purely learning-based methods or kinematic models with mismatched geometric trailer parameters.

The remainder of this paper is structured as follows.
Section~\ref{sec:approach} presents the approaches adopted in this work, including the kinematic modeling of articulated truck-semitrailer combinations, learning-based models, and the hybrid model under the \ac{ekf} framework.
Section~\ref{sec:vehicle} describes the test vehicle, as well as the data collection and preprocessing procedures.
Section~\ref{sec:experiments} presents the experimental setup and results.
Finally, conclusions are drawn in Section~\ref{sec:conclusion}.


\section{Methods}
\label{sec:approach}
This section first introduces the kinematic modeling of the truck–semitrailer combination. It then presents several learning-based models, including a monocular vision model with single-frame input, a model using single-frame kinematic and visual inputs, and a model using temporal sequential  kinematic and visual inputs. Finally, a hybrid model based on the \ac{ekf} is described, together with a method for dynamically adjusting the observation noise.

\subsection{Kinematic Model of the Truck-Semitrailer Combination}
To avoid complex parameter identification, a longitudinal kinematic model of the truck–semitrailer system is established solely for the estimation of the articulation angle. 
The longitudinal kinematics of the truck-semitrailer combination are represented by
\begin{equation}
\label{eq:kinematic}
	-\dot{\psi }+\frac{l^{\prime}}{l}\dot{\psi }\cos \theta +\frac{v}{l}\sin \theta+ \dot{\theta } = 0
\end{equation}
in which $\theta$ is the articulation angle between truck and semitrailer, $\dot{\psi }$ is the yaw rate of the truck, $v$ is the longitudinal velocity of the truck, $l$ and $l^{\prime}$ are the length from kingpin to the middle axles of trailer and to the rear axle of truck. The kinematic assumption requires a low lateral velocity of truck and semitrailer.

\subsection{Learning-based Model with Visual Input}
For the image input $\mathbf{I}_t \in \mathbb{R}^{224 \times 224 \times 3}$, a CNN-based regression model \emph{ResNet18/50-ArticNet} is developed.
Owing to its strong representation capability, stable optimization behavior enabled by residual connections, the model employs a ResNet-18/50 \cite{he2016deep} encoder as a feature extractor $\mathcal{E}_{\text{ResNet18}}(\cdot)$ or $\mathcal{E}_{\text{ResNet50}}(\cdot)$ to get feature maps $\mathbf{F}_{\text{I}_t} \in \mathbb{R}^{h \times w \times c}$ ($w$,
$h$ and $c$ denote the width, height, and number of channels of the feature map, respectively) as
\begin{equation}
\label{eq:encode}
 \mathbf{F}_\text{I} = \mathcal{E}_{\text{ResNet18}}( \mathbf{I} )
\end{equation}
followed by a global average pooling (GAP) for each channel to get feature vector $\boldsymbol{f}_{\text{I}_t} \in \mathbb{R}^{1 \times c}$.
Finally, a \ac{mlp} is employed to predict the articulation angle
\begin{equation}
\label{eq:MLP}
	\theta_t = ( \mathrm{ReLU}(\boldsymbol{f}_{\text{I}_t}\boldsymbol{W}_{\text{hidden}}))\boldsymbol{W}_{\text{final}}.
\end{equation}

For clarity, the ResNet18/50-ArticNet are referred to as ResNet18 and ResNet50 in the following, as only the task-specific decision layer is modified.
\subsection{Learning-based Model with Visual and Kinematic Input}
To combine both kinematic and visual information, a multimodal regression model is developed, referred to as \ac{vkit}, based on a cross-attention mechanism 
\cite{vaswani2017attention}. 
\begin{figure*}[t]
 \vspace{5pt}
  \centering
  \includegraphics[width=\textwidth]{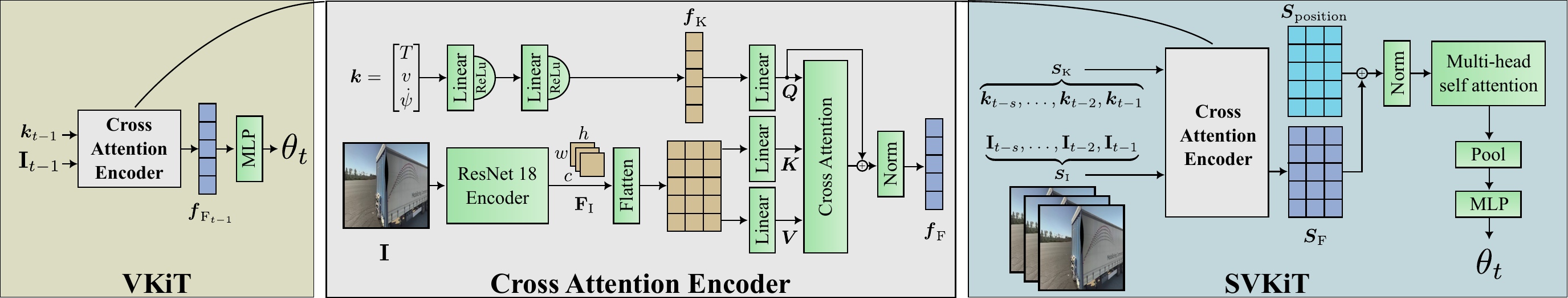}
  
  \caption{Overall framework of \ac{vkit} and \acs{svkit}. Both methods share the same cross attention encoder module.}
  \label{fig:VKiT}
   \vspace*{-5mm}
\end{figure*}
As shown in Figure~\ref{fig:VKiT} left, the model consists of a multimodal encoder and an \ac{mlp} in eq.~\ref{eq:MLP} as regressor. 

The image input $\mathbf{I}_{t-1}$ is processed by a ResNet18 encoder $\mathcal{E}_{\text{ResNet18}}(\cdot)$ to extract the feature maps $\mathbf{F}_{\text{I}_{t-1}} \in \mathbb{R}^{h \times w \times c}$, then it is flattened into a sequence of tokens $\hat{\boldsymbol{F}}_{\text{I}_{t-1}} \in \mathbb{R}^{hw \times c}$. The kinematic input vector $\boldsymbol{k}_{t-1} = \begin{pmatrix}v_{t-1}&\dot{ \boldsymbol{\psi}}_{t-1}& T_{t-1}\end{pmatrix}$
passes through two linear layers $\boldsymbol{W}_{\text{kin}_1}$ and $\boldsymbol{W}_{\text{kin}_2}$ with ReLU activation to obtain a kinematic feature embedding, in which $T$ is the time interval between two consecutive time steps.
\begin{equation}
	\boldsymbol{f}_{\text{K}_{t-1}}= \mathrm{ReLU}\left(\mathrm{ReLU}( \boldsymbol{k}_{t-1}\boldsymbol{W}_{\text{kin}_1})\boldsymbol{W}_{\text{kin}_2} \right)\in \mathbb{R}^{1 \times c^{\prime}}.
\end{equation}

The embedding $\boldsymbol{f}_{\text{K}_{t-1}}$ is used to obtain the \textit{query} $\boldsymbol{Q}\in \mathbb{R}^{1 \times d}$, while $\hat{\boldsymbol{F}}_{\text{I}_{t-1}}$ is utilized to obtain the \textit{key} $\boldsymbol{K}\in \mathbb{R}^{hw \times d}$ and \textit{value} $\boldsymbol{V}\in \mathbb{R}^{hw \times d}$ in a cross-attention module:
\begin{equation}
\begin{aligned}
        \boldsymbol{Q} = \boldsymbol{f}_{\text{K}}\boldsymbol{W}_{\text{Q}},\\
        \boldsymbol{K} = \hat{\boldsymbol{F}}_{\text{I}}\boldsymbol{W}_{\text{K}},\\
        \boldsymbol{V} = \hat{\boldsymbol{F}}_{\text{I}}\boldsymbol{W}_{\text{V}}.
	\end{aligned}
\end{equation}
The fused representation is then combined with $\boldsymbol{Q}$ via a residual connection to obtain the final multimodal feature
\begin{equation}
	\label{eq:CrossAttention}
		\boldsymbol{f}_\text{F} = \mathrm{LayerNorm}\left(\mathrm{softmax}\left(\frac{\boldsymbol{Q} \boldsymbol{K}^\top}{\sqrt{d}}\right) \boldsymbol{V} + \boldsymbol{Q}\right) \in \mathbb{R}^{1 \times d}.
\end{equation}
 The whole pipeline of embedding and fusion is summarized as the cross-attention-encoder $\mathcal{E}_{\text{VKiT}}(\cdot)$ (see Fig. \ref{fig:VKiT} middle) as
 \begin{equation}
\boldsymbol{f}_{\text{F}_{t-1}} = \mathcal{E}_{\text{VKiT}}(\mathbf{I}_{t-1},\boldsymbol{k}_{t-1}).
 \end{equation}
The fused feature vector $\boldsymbol{f}_{\text{F}_{t-1}}$ is passed through the MLP as defined in eq.~\ref{eq:MLP} with $\boldsymbol{f} = \boldsymbol{f}_{\text{F}_{t-1}}$ to predict the articulation angle $\theta_t$.
\subsection{Learning-based Model with Sequential Visual and Kinematic Input}

Since only the longitudinal kinematic vectors are available as input, the lateral kinematic information is missing.
In principle, the lateral kinematics can be inferred from a dynamics model.
To provide the model with information about the system's past behavior and implicitly guide it to learn the dynamics and lateral kinematics internally, the \ac{vkit} model is extended to handle sequential inputs, resulting in the \ac{svkit} model (see Fig. \ref{fig:VKiT} right).

Specifically, a sliding time window approach is adopted, feeding the model with a sequence of past $s$ discrete time steps of kinematic vectors, denoted as $\mathbf{S}_{\text{K}_t}=\left\{\boldsymbol{k}_{t-1}, \boldsymbol{k}_{t-2}, \dots, \boldsymbol{k}_{t-s}\right\}$, along with the corresponding sequence of images, denoted as $\mathbf{S}_{\text{I}_t}=\left\{\mathbf{I}_{t-1}, \mathbf{I}_{t-2}, \dots, \mathbf{I}_{t-s}\right\}$. This allows the model to leverage temporal information to better estimate the articulation angle.

The \ac{svkit} model employs the same cross-attention-encoder $\mathcal{E}_{\text{VKiT}}(\cdot)$ as \ac{vkit} to encode the historical sequence of $s$ discrete time steps, resulting in $s$ fused feature vectors $\boldsymbol{S}_{\text{F}_t}=\begin{pmatrix}\boldsymbol{f}_{\text{F}_{t-1}}^\top& \boldsymbol{f}_{\text{F}_{t-2}}^\top& \dots& \boldsymbol{f}_{\text{F}_{t-s}}^\top\end{pmatrix}^\top \in \mathbb{R}^{s \times d}$. These vectors are treated as $s$ tokens and are added to learnable positional embeddings $\boldsymbol{S}_{\text{position}} \in \mathbb{R}^{s \times d}$ \cite{dosovitskiy2020image}. The sequence of tokens is then processed by two multi-head self-attention blocks with residual connection and layer normalization $\mathcal{A}(\cdot)$ \cite{vaswani2017attention} 
to integrate information across the temporal dimension as sequence of historical information
\begin{equation}
	\label{eq:SVKIT1}
\boldsymbol{S}_{\text{history}} = \mathcal{A}\left(\mathcal{A}(\mathrm{LayerNorm} ( \boldsymbol{S}_{\text{F}} + \boldsymbol{S}_{\text{position}}))\right)\in \mathbb{R}^{s \times d}.
\end{equation}

Finally, the outputs are average-pooled \cite{dosovitskiy2020image} to obtain a fused representation of the historical information $\boldsymbol{f}_{\text{S}_t} \in \mathbb{R}^{1 \times d}$, which is passed through the same \ac{mlp} regressor as defined in eq.~\ref{eq:MLP} with $\boldsymbol{f} = \boldsymbol{f}_{\text{S}_t}$ to predict the articulation angle $\theta_t$.

\subsection{Hybrid Model based on Extended Kalman Filter}
The hybrid model allows the kinematic model to provide physical constraints and prior knowledge to the learning-based model, while the learning-based model can supply an initial state estimate to the kinematic model.
By leveraging the strengths of both approaches, the hybrid model can achieve more accurate and robust state estimation.
An \ac{ekf} is employed to integrate learning-based models ResNet18 and \ac{vkit} as measurement with the kinematic model defined in eq.~\ref{eq:kinematic} as the prediction.
They are referred to as \emph{EKF-ResNet18} and \emph{EKF-VKiT}, respectively.
The initial articulation angle $\theta_0$ is estimated by the visual (observation) model.
The prediction of the \ac{ekf} with state space $x_t = \theta_t  $ and control inputs $\boldsymbol{u}_t = \begin{pmatrix}\dot{\psi _t}&  v_t\end{pmatrix}^\top$ is formulated as follows:
\begin{equation}
\label{eq:ekf_p2}
 \hat{x}^-_t = g( \hat{x}_{t-1},\boldsymbol{u}_{t-1}),
\end{equation}
\begin{equation}
\label{eq:ekf_p3}
G_t = 
\left.
\frac{\partial g}{\partial x}
\right|_{ \hat{x}_{t-1}, \boldsymbol{u}_{t-1}},
\end{equation}

\begin{equation}
\label{eq:ekf_p4}
 P^-_t = G_{t} P_{t-1} G_{t}^\top+Q.
\end{equation}
Here, $Q$ is the noise covariance of state transition function $g(\cdot)$, $P$ is covariance of state $x$, $G$ is derivative of $g(\cdot)$, $\hat{x}$ is the estimated state. $\Box^-$ denote quantities in the prediction step.

The update with observation $\boldsymbol{z}_t$, which is estimated by a learning-based model, is formulated as follows:


\begin{equation}
\label{eq:ekf_o2}
H_t = 
\left.
\frac{\partial h}{\partial x}
\right|_{\hat{x}^-_{t}},
\end{equation}

\begin{equation}
\label{eq:ekf_o3}
K_t = P^-_tH_t^\top(H_tP^-_tH_t^\top+R_t)^{-1},
\end{equation}

\begin{equation}
\label{eq:ekf_o4}
\hat{x}_t = \hat{x}^-_t + K_t(\boldsymbol{z}_t-h(\hat{x}^-_t)),
\end{equation}

\begin{equation}
\label{eq:ekf_o5}
P_t = P^-_t - K_t H_tP^-_t.
\end{equation}
Here, $K$ is the Kalman gain, $H$ is the derivative of the observation function $h(\cdot)$, and $R$ is the measurement noise covariance.

\subsection{Soft-Measurement}
The performance and generalization capability of learning-based models are highly sensitive to the quality and diversity of the training dataset. Environmental factors such as lighting conditions, weather variations, and differences in vehicle appearance can introduce significant noise, thereby degrading model accuracy. In addition, the kinematic model itself can suffer from significant prediction errors when the distance $l$ from king-pin to trailer rear axles  is inaccurately specified, which is often the case in practical scenarios involving different trailers. Under such conditions, assuming a fixed observation noise covariance $R$ is insufficient, as it cannot accommodate the varying reliability of the measurement model across operating conditions.
Therefore, creating a dynamic noise covariance $R_t$ is essential to balance the contributions of the learning-based and kinematic models. To address this issue, the soft-measurement method in \cite{ewering2024reliable} is extended to the measurement involving visual inputs (\ac{vkit} and ResNet18). ResNet50 and \ac{svkit} are not suitable for the Soft Measurement method due to issues with the dimensionality of the input embedding representations.
Firstly, feature vectors $\boldsymbol{f}_\text{I}$ or $\boldsymbol{f}_\text{F}$ for every sample in the training set with $n$ samples are extracted and subjected to \ac{pca} to reduce their dimensionality 
to $\boldsymbol{f}_{\text{I, re}} \in \mathbb{R}^{1 \times r}$ or $\boldsymbol{f}_{\text{F, re}} \in \mathbb{R}^{1 \times r}$ with $ r<d$ and $r<c$. The reduced feature vectors $\boldsymbol{f}_{\text{I, re}}$ or $\boldsymbol{f}_{\text{F, re}}$ and PCA transformations $\mathcal{P}_{\text{resnet18}}(\cdot)$ or $\mathcal{P}_{\text{vkit}}(\cdot)$ are stored offline. 

During online inference, feature vectors $\hat{\boldsymbol{f}_\text{I}}$ or $\hat{\boldsymbol{f}_\text{F}}$ are derived from the input signals and transformed using the previously computed PCA parameters:
\begin{equation}
\label{eq:PCAreduceInference}
\begin{aligned}
  \hat{\boldsymbol{f}}_{\text{I, re}} = \mathcal{P}_{\text{resnet18}}( \hat{\boldsymbol{f}_\text{I}} ),\\
  \hat{\boldsymbol{f}}_{\text{F, re}} = \mathcal{P}_{\text{vkit}}(  \hat{\boldsymbol{f}_\text{F}} ).
\end{aligned}
\end{equation}
$\hat{\boldsymbol{f}}_{\text{I, re}}$ or $\hat{\boldsymbol{f}}_{\text{F, re}}$ are utilized to estimate the confidence of the model's prediction by comparing $\hat{\boldsymbol{f}}_{\text{I, re}}$ or $\hat{\boldsymbol{f}}_{\text{F, re}}$ to the distribution of training samples in the reduced feature space $\{  \boldsymbol{f}_{\text{I, re}_i} \} _{i=1}^{n}$ or $\{  \boldsymbol{f}_{\text{F, re}_ i} \} _{i=1}^{n}$.
The $L_2$ distances $\{  d_{\text{D}_i}  \} _{i=1}^{n}$ between $\hat{\boldsymbol{f}}_{\text{I, re}}$ and every $\boldsymbol{f}_{\text{I, re}_i}$ or between $\hat{\boldsymbol{f}}_{\text{F, re}}$ and every $\boldsymbol{f}_{\text{F, re}_i}$ are calculated, then the $k$ nearest distances $\{  d_{\text{d}_j} \} _{j=1}^{k} $ are found via 
\begin{equation}
\label{eq:knn}
d_{\text{k}_k} = \frac{1}{k} \sum_{j=1}^{k} d_{\text{d}_j}
\end{equation}
to calculate the noise covariance as
\begin{equation}
\label{eq:R}
R = (1+\gamma(\tau-1)^2)R_0
\end{equation}
with
\begin{equation}
\label{eq:tau}
\tau = 
\begin{cases}
\frac{d_{\text{max}}-d_{\text{k}_k}}{d_{\text{max}}} , & \text{if }  d_{\text{max}}>d_{\text{k}_k}\\
0, & \text{otherwise}.
\end{cases}
\end{equation}
$R_0$ and $\gamma$ are hyperparameters to adjust the scale, $d_{\max}$ is the threshold of average of $k$ nearst distances $d_{\text{k}_k}$, which determines under which conditions the model becomes completely unreliable.

The process noise covariance $Q$, the nominal measurement noise covariance $R_0$, and the confidence scaling factor $\gamma$ are empirically tuned on the calibration scenario.

During calibration, the confidence value $\tau$ is first fixed to $\tau = 1$, such that the observation noise reduces to a constant $R = R_0$. Under this condition, $Q$ and $R_0$ are tuned such that the EKF output closely follows the observation curve, effectively aligning the EKF estimates with the measurements.
Subsequently, the confidence is set to $\tau = 0$, corresponding to the lowest observation reliability. The parameter $\gamma$ is then adjusted to increase the observation noise $R$, ensuring that the EKF primarily relies on the prediction model in low-confidence situations. 

\section{Test vehicle and dataset}
\label{sec:vehicle}
\begin{figure}[b]
    \centering
    \def\svgwidth{0.9\columnwidth}
    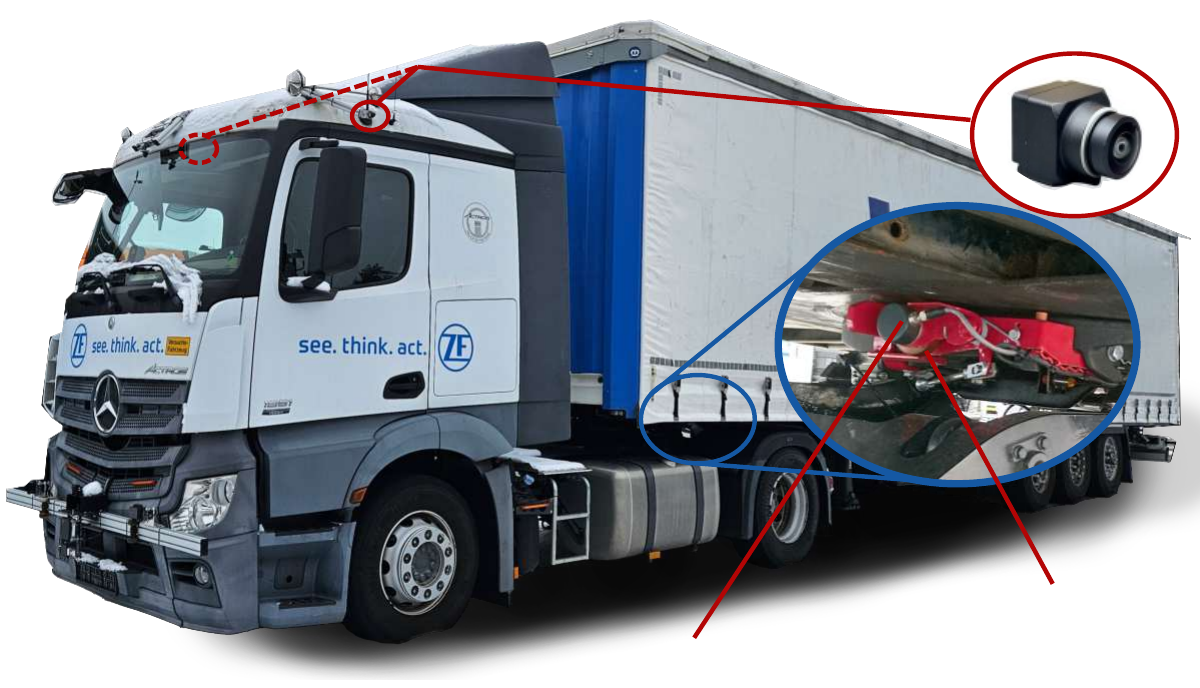
    \caption{The test vehicle used for image and ground truth dataset collection.}
    \label{fig:vehicle}
\end{figure}

\begin{figure}[t]
    \vspace{3pt}
    \centering
    \includegraphics[width=\columnwidth]{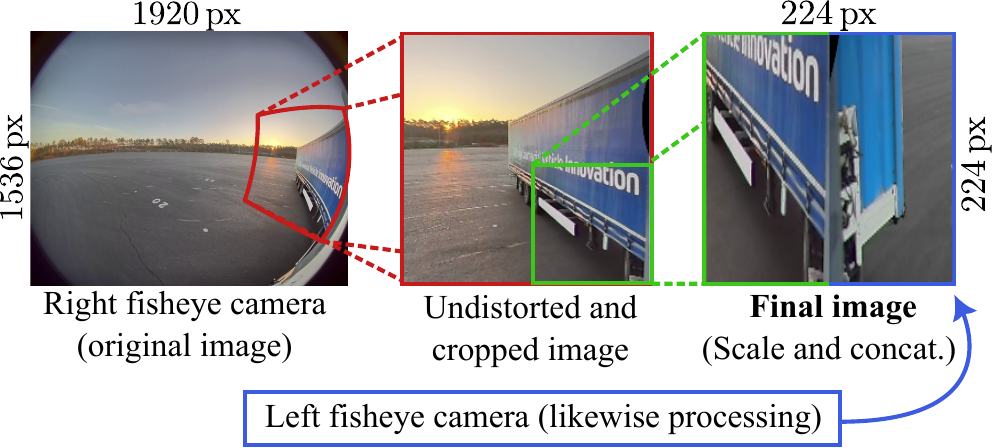}
    \caption{Processing steps of the fisheye camera images.
    The relevant region of each input image is first undistorted using the calibrated Kannala-Brandt intrinsics model \cite{kannala2006generic}.
    From the undistored output, the lower area of the image on the vehicle-facing side is extracted and scaled to \SI{50}{\percent} of its original width.
   The left and right views are then concatenated to form a single image of size $224\,\text{px} \times 224\,\text{px}$.}
    \label{fig:image_crop}
    \vspace{-5mm}
\end{figure}

The experimental setup comprises a Daimler Actros semi-truck (see Fig.~\ref{fig:vehicle}) coupled with four distinct trailers differing in geometry and dimensions.
Image data is recorded using two fisheye cameras (\emph{Sensing SG3S-ISX031C}, $1920\times1536\,\text{px}$ resolution, $30\,\text{fps}$, \SI{195}{\degree} horz. field-of-view).
The cameras are mounted above the nearside and offside side mirrors, with their optical axes oriented perpendicularly to the corresponding vehicle sides.
These locations correspond to typical mounting positions for camera‑mirror systems, providing an unobstructed view of the trailer and its surroundings.
While the fisheye cameras are not standard mirror‑replacement units, a side‑mirror‑like perspective is recovered through geometric undistortion of the captured images (see Fig.~\ref{fig:image_crop}).

The ground-truth articulation angle corresponding to the recorded trailer pose images is measured using a rotary encoder.
The encoder, fitted with a rubber contact wheel mounted near the fifth wheel, measures the trailer’s rotation by contacting its lower surface.
To correct signal offsets after trailer attachment and deviations caused by wheel slippage, the encoder output is compensated using a kinematic model via an \ac{ekf}.
This method was validated on different stationary scenarios in the relevant articulation angle range.
SAE $\text{J}1939$ CAN data (including vehicle speed, yaw rate, \etc) and articulation-angle sensor measurements are recorded at \SI{10}{\hertz}.
Image data is aquired at \SI{30}{\hertz} but synchronized and down-sampled to match the lower signal rate during post-processing.
The datasets are recorded on a test track featuring open spaces, parking lots, and highway-like segments.
\section{Experiments}
In this section, a series of experiments are designed to test and evaluate the approaches presented in Section~\ref{sec:approach}. The results of experiments are then presented.
\label{sec:experiments}
\subsection{Experimental Setup}
To evaluate model performance introduced in Sec.~\ref{sec:approach}, three experiments are designed targeting different aspects of generalization. 
Experiment I investigates robustness to variations in trailer color and pattern. 
Experiment II evaluates performance under different lighting conditions. 
Experiment III tests generalization to previously unseen trailer types. 
Table~\ref{tab:exp_overview} summarizes the experimental settings for each case and exhibits a visual overview of each dataset.
The solely learning-based models, 
namely ResNet18, ResNet50, \ac{vkit}, and \ac{svkit}, 
as well as the hybrid approaches EKF-ResNet18 and EKF-VKiT, 
are evaluated across all three experimental settings. For all three experiments, the hybrid models EKF-ResNet18 and EKF-VKiT 
employ the default geometric parameter $l$ of trailer T1 and T2 in the kinematic model. 
This setting is intended to simulate a realistic deployment scenario, 
where rapid adaptation to different trailers is required without prior 
recalibration of model parameters. 
In Experiment III, where trailer T4 is used for testing, the employed 
default parameter $l$ corresponds to 1.58 times its true geometric value. 
This introduces a substantial model mismatch in the kinematic component.

\begin{table}[t]
\centering
\caption{Overview of Experimental Settings}
\label{tab:exp_overview}
\setlength{\tabcolsep}{2pt}
\begin{tabular}{p{1.15cm}p{1.15cm}c|p{1.2cm}c|p{1.2cm}}
\toprule
ID & \multicolumn{2}{c|}{Train Set} & \multicolumn{2}{c|}{Test Set} & \# samples (Train/Test) \\
\midrule
I & 
T1; Sunny & 
\raisebox{-1.3cm}{\includegraphics[width=1.5cm]{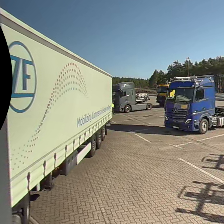}} & 
T2; Light rain & 
\raisebox{-1.3cm}{\includegraphics[width=1.5cm]{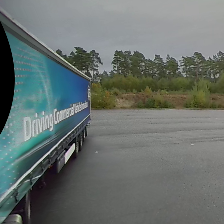}} & 
$8,996$ / $10,420$ \\[2pt]

II & 
T2; Light rain & 
\raisebox{-1.3cm}{\includegraphics[width=1.5cm]{Abb/ExampleSamples/T1b.png}} & 
T2; Night/Dawn & 
\raisebox{-1.3cm}{\includegraphics[width=1.5cm]{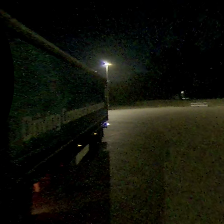}} & 
$8,336$ / $25,071$ \\[2pt]

III & 
T1 (as in I), T3; Sunny & 
\raisebox{-1.3cm}{\includegraphics[width=1.5cm]{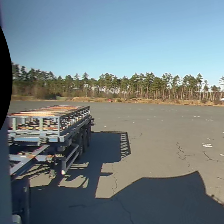}} & 
T4; Sunny & 
\raisebox{-1.3cm}{\includegraphics[width=1.5cm]{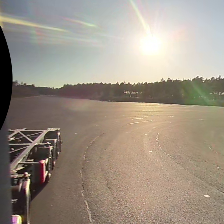}} & 
$20,679$ / $12,677$ \\[2pt]

\midrule
\multicolumn{6}{c}{Trailer Configuration} \\
\midrule
Trailer ID & \multicolumn{2}{l}{Trailer Type} & & & $l$ \\
\midrule
T1 & \multicolumn{2}{l}{Box Trailer (white)}   & & & 7.6m \\
T2 & \multicolumn{2}{l}{Box Trailer (blue)} & & & 7.6m \\
T3 & \multicolumn{2}{l}{Skeletal Trailer} & & & 7.5m \\
T4 & \multicolumn{2}{l}{Flatbed Trailer}  & & & 4.8m \\
\bottomrule
\end{tabular}
\vspace{-6.3mm}
\end{table}

The learning-based models are trained based on MSE loss function utilizing the AdamW optimizer \cite{loshchilov2017decoupled} with weight decay.
Online data augmentations for images are employed during training, including random color jittering, random gamma adjustment, random Gaussian noise, and random erasing.
The dataset is randomly split into 80\% training and 20\% validation sets.
The distributions of ground truth in both subsets are approximately the same.
The training employs early stopping based on validation loss, with empirically optimized hyperparameters for stable convergence.
For the \ac{svkit} model, a learning rate warmup is also employed to stabilize self-attention training \cite{vaswani2017attention}.

In Experiments I and II, the encoder structures of all learning-based models 
(ResNet18 and ResNet50 backbones) are initialized with ImageNet-1k-pretrained weights \cite{paszke2019pytorch}
and subsequently fine-tuned on the target dataset to accelerate convergence.
For Experiment III, a staged fine-tuning strategy is adopted. The models trained in Experiment I on the T1 trailer are used for initialization. 
Based on this initialization, the networks are further fine-tuned on a mixed 
dataset consisting of T1 and T3 trailers to enhance adaptation capability 
to unseen trailer types.

\subsection{Results of Experiments}

Given the substantial imbalance in the distribution of articulation angles during data collection, with limited samples exceeding ±45° in some datasets,  the quantitative results presented in Table~\ref{tab:comparison} 
are evaluated exclusively on test samples within the ±45° range 
for all three experiments.

\begin{figure}
        \begin{center}
            \includegraphics[width=0.98\columnwidth]{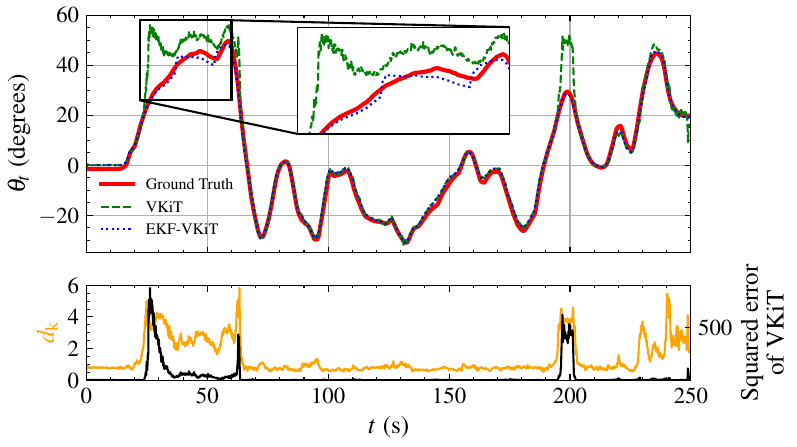}
            \vspace{-2mm}
            \caption[Representative scenario trajectory of the EKF-VKiT model in experiment I]{Scenario trajectories of the EKF-VKiT model in experiment I. In this scenario, the VKiT model exhibits a relatively high number of outlier estimation errors, which are associated with increased $d_\text{k}$ values. The EKF effectively mitigates these outliers through the fusion of model-based predictions and visual measurements. }
            \label{fig:vkit_I}
        \end{center}
        \begin{center}
            \includegraphics[width=0.98\columnwidth]{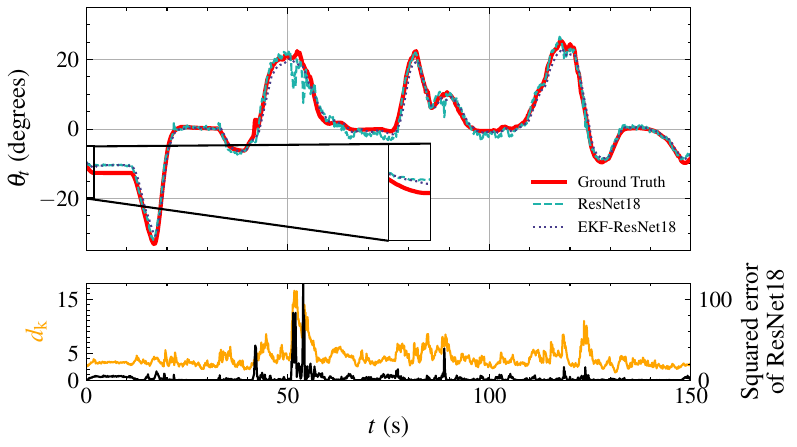}
            \vspace{-2mm}
            \caption{Scenario trajectories of the EKF-ResNet18 model in Experiment III. Shown are ResNet18 and EKF-ResNet18 as examples; all models provide accurate initial angle estimates even for arbitrary starting angles.}
            \label{fig:resnet_init}
        \end{center}
        \vspace{-5mm}
\end{figure}

It can be observed that all models except SVKiT and VKiT achieve strong performance in experiment~I, with the \ac{mae} being close to $1.0^\circ$, VKiT shows slightly higher \ac{mae} and markedly higher \ac{rmse}, which is primarily attributed to multiple outliers occurring in a specific scenario, as illustrated in Fig.~\ref{fig:vkit_I}, where samples with large prediction errors in the \ac{vkit} are associated with higher estimated uncertainty, which is represented by the k-NN distance $d_{\text{k}}$ 
defined in eq.~\ref{eq:knn}.
Within the hybrid framework, these high-uncertainty observations are down-weighted, causing the state estimate to rely more heavily on the kinematic model, thereby improving overall robustness and accuracy.
The hybrid model EKF-VKiT achieves significantly higher accuracy than the standalone learning-based observation model VKiT.

\begin{table}[t]
\centering
\caption{Performance Comparison}
\label{tab:comparison}
\setlength{\tabcolsep}{3pt}   

\begin{tabular}{l cc cc cc}
\toprule
\multirow{2}{*}{Models} 
& \multicolumn{2}{c}{Exp. I}
& \multicolumn{2}{c}{Exp. II}
& \multicolumn{2}{c}{Exp. III} \\
\cmidrule(lr){2-3}
\cmidrule(lr){4-5}
\cmidrule(lr){6-7}
& RMSE & MAE & RMSE & MAE & RMSE & MAE \\
\midrule
ResNet18        & $1.57^\circ$ & $1.16^\circ$ & $2.88^\circ$ & $2.06^\circ$ & \bm{$1.89^\circ$} & $1.39^\circ$ \\
ResNet50        & $1.04^\circ$ & \bm{$0.74^\circ$} & $4.41^\circ$ & $2.86^\circ$ & $3.08^\circ$ & $1.81^\circ$ \\
VKiT            & $3.09^\circ$ & $1.43^\circ$ & $3.04^\circ$ & $1.80^\circ$ & $2.04^\circ$ & \bm{$1.12^\circ$} \\
SVKiT           & $2.75^\circ$ & $2.06^\circ$ & $4.47^\circ$ & $3.07^\circ$ & $2.13^\circ$ & $1.43^\circ$ \\
EKF-ResNet18    & $1.19^\circ$ & $0.93^\circ$ & \bm{$1.46^\circ$} & \bm{$0.95^\circ$} & $2.46^\circ$ & $1.78^\circ$ \\
EKF-VKiT        & \bm{$1.02^\circ$} & $0.79^\circ$ & $2.69^\circ$ & $1.46^\circ$ & $2.64^\circ$ & $1.63^\circ$ \\
\bottomrule
\end{tabular}
 \vspace{-3mm}
\end{table}



In experiment~II, the test set corresponds to a night and dawn scenario with extremely low ambient illumination.
Streetlights are present, and their relative intensity and angles vary across samples, causing significant lighting changes. Table~\ref{tab:comparison} shows that the extreme illumination conditions in experiment II lead to a noticeable performance degradation in MAE and RMSE across all models. Among the solely learning-based models, VKiT achieves the best performance in terms of MAE, however, its RMSE is slightly higher than that of ResNet18, indicating the presence of more pronounced outliers. EKF-ResNet18 and EKF-VKiT demonstrate superior overall performance in both RMSE 
and MAE compared to the other models. Since ResNet18 and VKiT tend to produce 
higher uncertainty estimates when large prediction errors occur, the EKF is 
able to down-weight these unreliable observations. Consequently, the filter 
effectively suppresses the influence of outliers, leading to improved robustness 
and overall estimation accuracy.

Since image-based data augmentation is employed during model training, the higher-than-expected performance observed in experiment~II is investigated with respect to its potential attribution to the applied augmentation strategy. Notably, the training data in experiment~II were recorded exclusively during daytime. To this end, an ablation study is performed in which data augmentation is omitted during training.
In comparison to learning-based models trained without data augmentation, all learning-based models exhibit a pronounced performance gain in experiment II with approximately 60\%--80\% reductions in both MAE and RMSE.
In addition the estimated observation uncertainty for ResNet18 and VKiT  in both hybrid models exhibits an overall increase when data augmentation is not applied.

In Experiment III, most solely learning-based models achieve MAE values of approximately $1.5^\circ$. Exception is ResNet50, whose MAE value is noticeably higher, reaching $1.81^\circ$. Evaluations on different trailer types with significantly mismatched geometric parameters show a noticeable degradation in the performance of both hybrid models. However, due to the corrective mechanism of the hybrid models with soft-measurement, the overall estimation accuracy remains within an acceptable range, relative to the corresponding kinematic prediction models within the EKF, 
EKF-VKiT and EKF-ResNet18 achieve RMSE reductions of 59.9\% and 62.7\%, 
and MAE reductions of 60.7\% and 57.2\%, respectively.

Moreover, all models can handle scenarios with arbitrary initial articulation angles, without a straight-driving initialization. Fig.\ref{fig:resnet_init} illustrates a representative scenario in which the articulated truck starts with an articulation angle of approximately $-10^\circ$. ResNet18 and EKF-ResNet18 are shown as example models, both providing accurate initial angle estimates.

\section{Conclusion}
\label{sec:conclusion}
This paper presents learning-based and hybrid approaches for articulation angle estimation of truck–semitrailer systems using only signals from existing onboard truck sensors, eliminating the need for additional sensors or optical markers.
Models are trained and extensively evaluated on multiple \ac{ood} real-world datasets.
Hybrid models demonstrate high estimation accuracy and strong robustness under different challenging conditions.
Unlike conventional methods, it needs no straight-driving initialization and works at arbitrary initial angles.
Data augmentation during training enhances generalization, and the model is further evaluated under extremely low-light conditions.
Results show augmentation improves performance and reduces uncertainty.
Moreover, the proposed methods can be rapidly adapted to different trailer types. When trailer geometric parameters are unknown, predefined nominal values can be employed.
Even in cases where these predefined parameters deviate substantially from the true values (e.g., by a factor of 1.5), the hybrid model exhibits only moderate performance degradation and maintains robust estimation capability.
However, both the kinematic model and the learning-based model perform poorly on samples with large angles, so their complementarity for these samples is limited.
Moreover, the distance computations for a large number of samples in the soft-measurement approach incur high computational cost, which may affect real-time performance during online estimation.
Solely learning-based models perform frame-wise predictions without autoregressive propagation, avoiding temporal error accumulation. Therefore, the \ac{ekf} continuously corrects the kinematic prediction through the observation model, which is expected to alleviate long-term drift accumulation, although explicit long-term error propagation is not separately analyzed in this work.
In future work, the geometric parameters of the trailer could be incorporated into the \ac{ekf} state space, samples with lower observation uncertainty could then be utilized to dynamically update the nominal parameter values, thereby further improving adaptability to previously unseen trailers.



\bibliographystyle{IEEEtran}
\vspace{-0.5mm}
\bibliography{ref}

\end{document}

%% file: Abb/Components.pdf_tex
\begingroup%
  \makeatletter%
  \providecommand\color[2][]{%
    \errmessage{(Inkscape) Color is used for the text in Inkscape, but the package 'color.sty' is not loaded}%
    \renewcommand\color[2][]{}%
  }%
  \providecommand\transparent[1]{%
    \errmessage{(Inkscape) Transparency is used (non-zero) for the text in Inkscape, but the package 'transparent.sty' is not loaded}%
    \renewcommand\transparent[1]{}%
  }%
  \providecommand\rotatebox[2]{#2}%
  \newcommand*\fsize{\dimexpr\f@size pt\relax}%
  \newcommand*\lineheight[1]{\fontsize{\fsize}{#1\fsize}\selectfont}%
  \ifx\svgwidth\undefined%
    \setlength{\unitlength}{575.92157319bp}%
    \ifx\svgscale\undefined%
      \relax%
    \else%
      \setlength{\unitlength}{\unitlength * \real{\svgscale}}%
    \fi%
  \else%
    \setlength{\unitlength}{\svgwidth}%
  \fi%
  \global\let\svgwidth\undefined%
  \global\let\svgscale\undefined%
  \makeatother%
  \begin{picture}(1,0.57277109)%
    \lineheight{1}%
    \setlength\tabcolsep{0pt}%
    \put(0,0){\includegraphics[width=\unitlength,page=1]{Components.pdf}}%
    \put(0.68118142,0.5411454){\makebox(0,0)[lt]{\lineheight{1.25}\smash{\begin{tabular}[t]{l}Fisheye cameras\end{tabular}}}}%
    \put(0.50655469,0.00983021){\makebox(0,0)[lt]{\lineheight{1.25}\smash{\begin{tabular}[t]{l}Incremental rotary encoder\end{tabular}}}}%
    \put(0.74670535,0.05460487){\makebox(0,0)[lt]{\lineheight{1.25}\smash{\begin{tabular}[t]{l}Rubber wheel\end{tabular}}}}%
  \end{picture}%
\endgroup%